\documentclass[letterpaper, 10 pt, conference]{ieeeconf}  
\usepackage{amsmath,amsfonts}
\usepackage{algorithmic}
\usepackage{array}
\usepackage[caption=false,font=normalsize,labelfont=sf,textfont=sf]{subfig}
\usepackage{textcomp}
\usepackage{stfloats}
\usepackage{url}
\usepackage{verbatim}
\usepackage{graphicx}
\usepackage{booktabs}
\usepackage{balance}
\usepackage{cite}
\usepackage{algorithm}
\usepackage{algorithmic}
\usepackage{diagbox}
\usepackage{multirow}
\usepackage{makecell}
\usepackage{bbm}

\def\pos{\boldsymbol{p}}
\def\vel{\boldsymbol{v}}

\def\euler{\boldsymbol{\Theta}}
\def\angvel{\boldsymbol{\omega}}

\def\force{\boldsymbol{f}}
\def\gravity{\boldsymbol{g}}
\def\torque{\boldsymbol{\tau}}

\def\heading{\boldsymbol{h}}
\def\los{\boldsymbol{l}}

\def\PIDcoe{\boldsymbol{k}}

\DeclareMathOperator{\clip}{clip}
\DeclareMathOperator{\PID}{PID}

\def\path{\boldsymbol{P}}

\IEEEoverridecommandlockouts                              

\title{\LARGE \bf
Interpretable DRL-based Maneuver Decision of UCAV Dogfight
}

\author{Haoran Han$^{1}$, Jian Cheng$^{1}$, and Maolong Lv$^{2}$
\thanks{$^{1}$Haoran Han and Jian Cheng are with the School of Information and Communication Engineering, University of Electronic Science and Technology of China,  Chengdu 611731, China. Email: hanadam@163.com; chengjian@uestc.edu.cn}
\thanks{$^{2}$Maolong Lv is with the Air Traffic Control and Navigation College, Air Force
	Engineering University, Xi'an 710051, China, and also with College of
	Aeronautics Engineering, Air Force Engineering University, Xi'an 710038,
	China. Email: maolonglv@163.com}
}

\begin{document}

\maketitle
\thispagestyle{empty}
\pagestyle{empty}

\begin{abstract}

This paper proposes a three-layer unmanned combat aerial vehicle (UCAV) dogfight frame where Deep reinforcement learning (DRL) is responsible for high-level maneuver decision. A four-channel low-level control law is firstly constructed, followed by a library containing eight basic flight maneuvers (BFMs). Double deep Q network (DDQN) is applied for BFM selection in UCAV dogfight, where the opponent strategy during the training process is constructed with DT. Our simulation result shows that, the agent can achieve a win rate of 85.75\% against the DT strategy, and positive results when facing various unseen opponents. Based on the proposed frame, interpretability of the DRL-based dogfight is significantly improved. The agent performs yo-yo to adjust its turn rate and gain higher maneuverability. Emergence of "Dive and Chase" behavior also indicates the agent can generate a novel tactic that utilizes the drawback of its opponent.
\end{abstract}

\section{Introduction}

Over the last few years, unmanned combat air vehicle (UCAV) has become increasingly important in the battlefield. Most conventional aerial combat methods, such as decision tree (DT) \cite{DT} or state machine \cite{stateMachine}, are easy to implement and interpret. This process can be enhanced with target maneuver recognition \cite{semiSupervised} technique. However, they are impossible to cover all combat situations due to the complicated nonlinear nature of aerial combat.
Therefore, more researchers focused on the decision of basic flight maneuver (BFM) with optimization techniques \cite{Pigeon1, Pigeon2, Pigeon3}.

However, the strong non-linearity of aerial combat cause significant difficulty for optimization. Therefore, deep reinforcement learning (DRL), a data-driven optimization technique, has been attached more attention. 
The first DRL algorithm cluster is to let the agent generate the control command directly. Pope et al. \cite{DARPA} proposed a hierarchical DRL dogfight frame, where a network selected three controllers with different combat styles. Chai et al. \cite{hierarchicalContinue} utilized two cascade DRL controllers to output the desired attitude and control command respectively. Moreover, techniques such as transformer \cite{transformerCommand} and inverse DRL \cite{inverseCommand} were also applied.

In addition, DRL are also utilized for maneuver decision. The first category of BFM library consists of turning with different directions and loads \cite{11missleBFM}. Another library design consists of meaningful BFM, such as tracking, turn and split-s. Sun et al. \cite{Sun} applied self-play technique and observed emergence of multiple interpretable tactics. Piao et al. \cite{Piao} combined DRL with expert knowledge, and Zhu et al. \cite{Zhu} designed an interruption mechanism where the agent increased decision frequency when facing emergency.

However, most works concerning DRL-based dogfight \cite{transformerCommand, inverseCommand, 11missleBFM, Sun, Piao, Zhu} only focused on simplified UCAV dynamics with three degrees of freedom (DOF), which is far from the real situation. 
Moreover, most works only claimed their algorithms outperformed conventional methods. Some simple reasons were given, such as "more aggressive" \cite{Pigeon1, DARPA, hierarchicalContinue, transformerCommand}, but a deep interpretation of how DRL agent performs is still lacking. In response to these problems, this paper aims to enhance the interpretability of DRL behaviors when facing dogfights under realistic 6-DOF dynamics. Therefore, this paper contributes to the field of dogfight in three aspects.

Firstly, this paper proposes a three-layer UCAV dogfight decision frame. We firstly construct a four-channel low-level control law, followed bt a library consisting of eight common-apply BFMs. A bank-to-turn tracking algorithm is adopted, and then generalized to other BFMs. Lastly, the double deep Q network (DDQN) \cite{DDQN} is used to achieve the maneuver decision. Our simulation shows that, the agent can achieve a success rate of 85.75\% against its DT opponent. 

Then, this paper gives an in-depth post-hoc interpretation on how the agent behaves, mainly in two aspects. (1) This paper observes that, the agent can use yo-yo to adjust its rate of turn, so that it can gain higher maneuverability during a double-loop dogfight. (2) Furthermore, we also observe the emergence of "Dive and Chase" strategy, where the agent firstly decrease its height, and then chase its opponent who climbs to avoid crashing into the ground. 

Lastly, this paper provides a gym-style environment for the DRL-based maneuver decision at
https://github.com/Han-Adam/Dogfight, where F16 \cite{F16} is adopted.

\section{Preliminaries}
\subsection{UCAV Dynamical Model}

This paper adopts north-west-down fame as the earth frame. The translation dynamics is considered as 
\begin{equation}
	\begin{matrix}
		\dot{\pos} = R_{\text{b}}^{\text{e}}\vel^{\text{b}}, 
		&\dot{\vel}^{\text{b}} = (\force^{\text{t}}-\force^{\text{a}})/m + R_{\text{e}}^{\text{b}}\gravity - \euler\vel^{\text{b}},
	\end{matrix}
\end{equation}
where $m$ is the mass, $\pos=[p_1, p_2, p_3]^T$ is the position, $\vel^{\text{b}}=[u,v,w]^T$ is the velocity under the body frame, $\gravity=[0,0,g]^T$ is the gravity, $\euler=[\phi,\theta,\psi]^T$ is the Euler angle, $R_{\text{b}}^{\text{e}}$ is the rotation matrix from body to earth frame, $R_{\text{e}}^{\text{b}}=R_{\text{b}}^{\text{e},T}$,
and $\force^\text{t}=[f, 0, 0]$ and $\force^\text{a}=[D, L, C]^T$ are the thrust and air force. Moreover, the rotation dynmaics is considered as
\begin{equation}
	\begin{matrix}
		\dot{\euler}=H(\euler)\angvel, & \dot{\angvel}=J^{-1}(\torque-\angvel\times J\angvel)
	\end{matrix}
\end{equation}
where $J$ is the inertia matrix, $H(\euler)$ is the matrix transforming the angular velocity $\angvel=[p,q,r]^T$ into the change rate of Euler angle, $\torque=[l,m,n]^T$ is the external torque. The ground speed $V$, attack angle $\alpha$, and sideslip angle $\beta$ are
\begin{equation}
	\begin{matrix}
		V=||\vel^{\text{b}}||, &\alpha=\arctan(w/u), &\beta=\arcsin(v/V).
	\end{matrix}
\end{equation}

Additionally, the mach number $Ma$ is defined as the ratio between the ground speed and the air speed. The force and the torque generated by the air effect is calculated as
\begin{equation}
	\begin{matrix}
		D=\bar{q}SC_D, &L=\bar{q}SC_d, &C=\bar{q}SC_C,\\
		l=\bar{q}SbC_l, &m=\bar{q}S\bar{c}C_m, &n=\bar{q}SbC_n,
	\end{matrix}
\end{equation}
where $\bar{q}$ is the dynamic press, $S$ and $b$ are the area and span for the wing, $\bar{c}$ is the reference length. The aerodynamic coefficients $C_\xi$, where $\xi=D,L,C,l,m,n$, are determined by the deflection of elevator $\delta^{\text{e}}$, aileron $\delta^{\text{a}}$, rudder $\delta^{\text{r}}$, $\alpha$, $\beta$, and $Ma$. In addition, the thrust $f$ is determined by the throttle $\delta^{\text{t}}$. According to \cite{F16}, the amplitude and the change rate of the control input $\boldsymbol{\delta}=[\delta^{\text{t}}, \delta^{\text{e}}, \delta^{\text{a}}, \delta^{\text{r}}]$ are all restricted.

\subsection{Simplified 3-DOF Model}
In the simplified model, the heading is equivalent to the direction of the velocity. Therefore, the 3-DOF model is
\begin{equation}
	\begin{matrix}
		\dot{p}_1 = V\cos\chi\cos\zeta, & \dot{p}_2 =V\cos\chi\sin\zeta, &
		\dot{p}_3 =-V\sin\chi,\\
		\dot{V}=g(n^\text{l}-\sin\chi), &
		\dot{\chi}=\frac{g(n^\text{n}\cos\phi-\cos\chi)}{V}, &
		\dot{\zeta}=\frac{gn^\text{n}\sin\phi}{V\cos\chi},
	\end{matrix}
\end{equation}
where $\chi$ and $\zeta$ represent the path pitch and path yaw, $n^\text{l}$ and $n^\text{n}$ represent the longitudinal and normal load. Here, the roll angle $\phi$, also known as bank angle, can be instantly changed.

\begin{figure}[!t]
	\centering
	\includegraphics[]{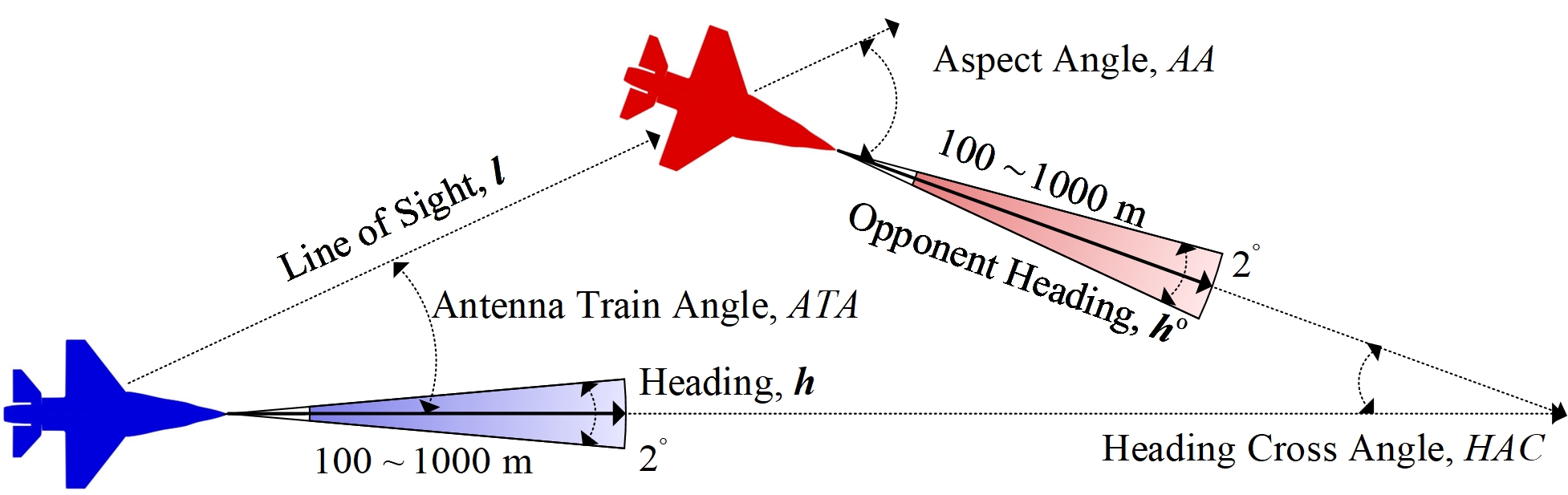}
	\caption{Geometric relation of UCAVs in the dogfight, where the blue one is controlled by the agent, and the red one is regarded as the opponent.}
\end{figure}

\subsection{Dogfight Setting}
the line of sight (LOS) is defined as $\los=\pos^{\text{o}}-\pos=[\Delta p_1, \Delta p_2, \Delta p_3]^T$, where the superscript of o represents the value of the opponent. The distance, heading crossing angle (HCA), antenna train angle (ATA), and aspect angle (AA) \cite{angleSetting} between a UCAV and its opponent are defined as
\begin{equation}
	\begin{matrix}
		d = ||\los||, & HCA=\arccos(\heading^T\heading^{\text{o}}), \\
		ATA=\arccos(\los^T\heading/d), &
		AA=\arccos(\los^T\heading^{\text{o}}/d),
	\end{matrix}
\end{equation}
where $\heading$ is the heading of the UCAV. The Geometic relation is illustrated in Fig. 1. The damage caused by weapons is constant, and the engagement zone [17] is designed as
\begin{equation}
	\begin{matrix}
		100\text{ m}\leq d \leq 1000\text{ m}, & ATA \leq 1^{\circ}.
	\end{matrix}
\end{equation}

The initial blood of each UCAV is assigned as one, which means it can afford damage for only one second. The UCAV is regarded as crashed, if its distance to the opponent is less than 10 m, or its height is less than 10 m. 

\subsection{Deep Reinforcement Learning}
The reinforcement learning is modeled as Markov decision process (MDP), $(\mathcal{S}, \mathcal{A}, \mathcal{R}, \mathcal{P}, \gamma)$. At each time $t$, the agent receives a state $s_{t}\in\mathcal{S}$ and takes an action $a_{t}\in\mathcal{A}$. The state of the agent then changes based on $s_{t+1}\sim\mathcal{P}(\cdot|s_t,a_t)$, and the instantaneous reward is generated based on $r_t=\mathcal{R}(s_t,s_{t+1})$. The discounted reward is defined as $G_t=\sum_{i=0}^{\infty}\gamma^{i}r_{t+i}$, where the discount factor $\gamma\in(0,1)$. The agent samples actions based on its policy $\pi$, and the Q function for a given policy is defined as 
$Q_\pi(s_t,a_t)=\mathbb{E}_{a_i\sim\pi(\cdot|s_i), s_{i+1}\sim\mathcal{P}(\cdot|s_{i},a_{i}),i>t}[G_t|s_t, a_t]$.

DRL uses neural networks to approximate the Q function. This paper uses $\eta$ and $\hat{\eta}$ to denote the parameter of real and target networks respectively. The policy is induced by 
\begin{equation}
	\label{epsilonGreedy}
	\pi(\cdot|s_t) = \left\{
	\begin{matrix}
		\arg\max(Q_\eta(s_t)) & \text{with probability of } \epsilon.\\
		U(\mathcal{A}) &  \text{with probability of } 1 - \epsilon.
	\end{matrix}\right.
\end{equation}
where $U(\cdot)$ is uniform distribution, and the greedy factor $\epsilon$ belongs to $(0, 1)$ during the train process, and equals to $1$ during the test process. This paper adopts DDQN to update the parameter, where the loss function is defined as
\begin{equation}
	\label{lossF}
	\begin{aligned}
		L(\eta) = \mathbb{E}_{\mathcal{D}} (&r_t + (1-\mathbb{I}(done_t)) \\
		& Q_{\hat{\eta}}(s_t, \max_{a}Q_{\eta}(s_{t+1},a)) - Q_{\eta}(s_t,a_t))^2.
	\end{aligned}
\end{equation}
Here, $\mathcal{D}$ is the replay buffer, from which all trajectories are uniformly sampled, $\mathbb{I}(\cdot)$ equals one if the statement is True or zero otherwise, $done_t$ represent whether the state is terminal.
The target network parameters $\hat{\eta}$ is updated for every $\widetilde{C}$ steps.

\section{Methodology}
A three-layer hierarchical dogfight frame, as illustrated in Fig. 2 is described in this section.

\begin{figure}[!t]
	\centering
	\includegraphics[]{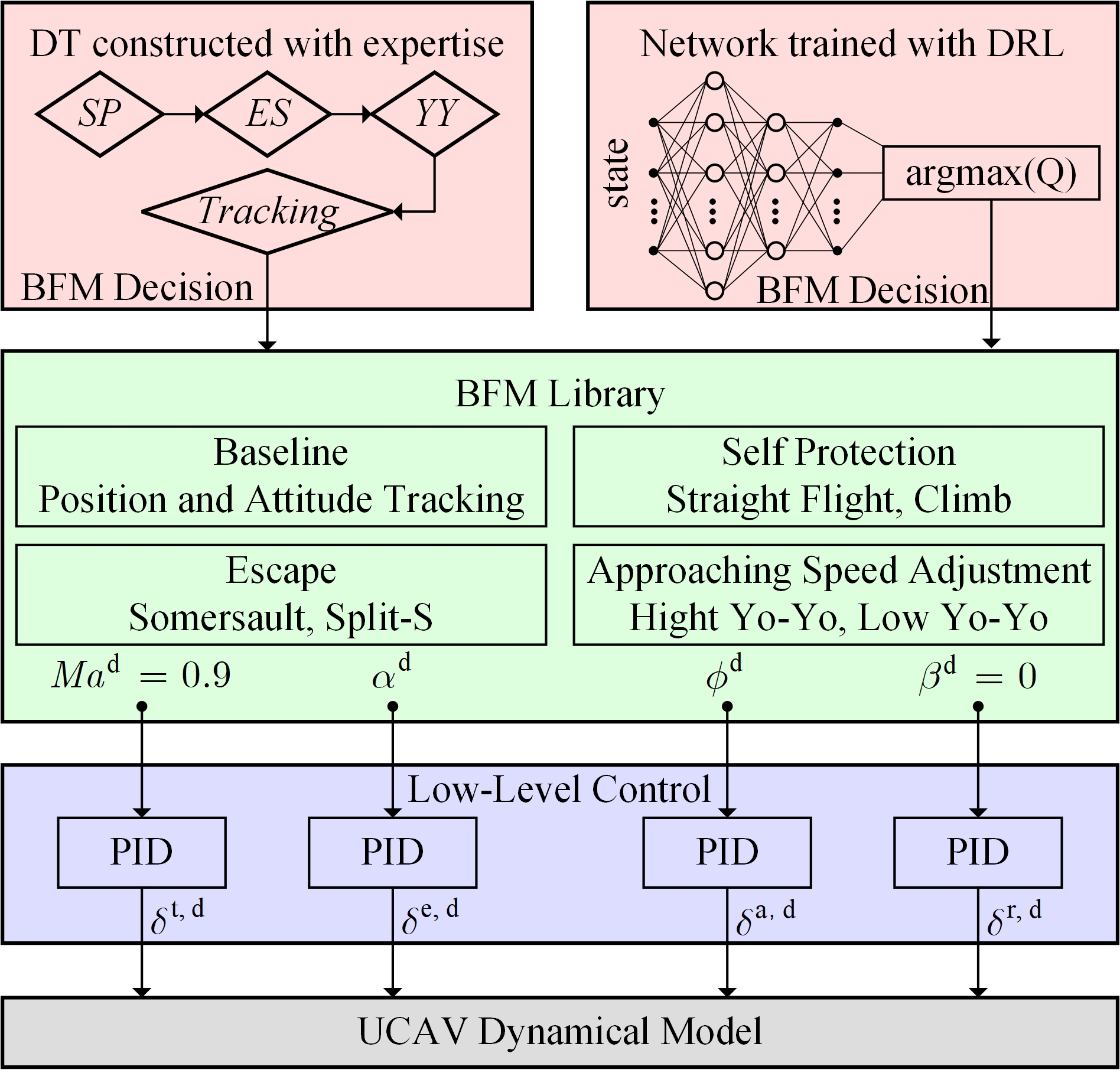}
	\caption{Overview of the proposed UCAV dogfight frame.}
\end{figure}

\subsection{Low-Level Control Law}
This paper uses widely-applied PID, which is defined as
\begin{equation}
	\PID(\xi, \xi^\text{d}, \PIDcoe_\xi)=k^\text{P}_\xi e_\xi + \int k^\text{I}_\xi e_\xi - k^\text{D}_\xi \dot{e}_\xi,
\end{equation}
for low-level control, where $\xi^\text{d}$ is the desired value of a variable $\xi$, $e_\xi=\xi^\text{d} - \xi$ is the error, and $\PIDcoe_\xi=[k^\text{P}_\xi, k^\text{I}_\xi, k^\text{D}_\xi]^T$. 

This paper divides the UCAV into four channels. First, the UCAV is set to chase a predefined mach number of 0.9, because a high velocity will cause performance degradation. Then, the elevator and aileron are adjusted based on the angle of attack $\alpha$ and roll $\phi$ respectively. This paper adopts bank-to-turn control mode. Thus, we directly use rudder to maintain $\beta$ at zero. 
In summary, the low level control law is
\begin{equation}
	\begin{aligned}
		\delta^\text{t, d} &= \clip(\PID(Ma,0.9,\PIDcoe_{Ma}),\delta^\text{t}_{\text{min}},\delta^\text{t}_{\text{max}}), \\
		\delta^\text{e, d} &= \clip(\PID(\alpha,\alpha^\text{d}, \PIDcoe_{\alpha}), \delta^\text{e}_{\text{min}}, \delta^\text{e}_{\text{max}}),\\
		\delta^\text{a, d} &= \clip(\PID(\phi,\phi^\text{d},\PIDcoe_{\phi}), \delta^\text{a}_{\text{min}}, \delta^\text{a}_{\text{max}}),\\
		\delta^\text{r, d} &= \clip(\PID(\beta, 0,\PIDcoe_{\beta}), \delta^\text{r}_{\text{min}}, \delta^\text{r}_{\text{max}}).
	\end{aligned}
\end{equation}

\subsection{Basic Flight Maneuver Library Construction}
The BFM library constructed in this paper consists of eight operations, namely position tracking, attitude tracking, straight flight, climb, somersault, split-s, high yo-yo and low yo-yo. They are built based on the simplified model.

\textit{Position Tracking}: The velocity vector of the UACV should point to the opponent. Therefore, the desired path pitch and path yaw are firstly calculated as
\begin{equation}
	\label{desiredAngValue}
	\begin{aligned}
		\chi^{\text{d}} &= \arctan(-\Delta p_3/\sqrt{\Delta p_1^2+\Delta p_2^2}),\\ 
		\zeta^{\text{d}} &= \arctan(\Delta p_2/ \Delta p_1).
	\end{aligned}
\end{equation}
Then, the load under the trajectory frame is calculated as
\begin{equation}
	\label{loadCalculation}
	\begin{matrix}
		n_2 = \frac{V}{g} k_\zeta e_\zeta \cos\chi, &
		n_3 = \frac{V}{g} k_\chi e_\chi + \cos\chi.
	\end{matrix}
\end{equation}
The desired roll angle and normal load are generated as
\begin{equation}
	\begin{matrix}
		\phi^\text{d}=\arctan(n_3/ n_2), & 
		n^\text{n} = \clip(n_2\sin\phi+n_3\cos\phi).
	\end{matrix}
\end{equation}

The normal load is transformed with 
\begin{equation}
	\alpha^\text{d}=\clip(k_n^\alpha n^{\text{n}}, \alpha^{\text{d}}_{\text{min}}, \alpha^{\text{d}}_{\text{max}})
\end{equation}
where $k_n^\alpha$ can be approximately set as a constant, and $\alpha^{\text{d}}_{\text{min}}, \alpha^{\text{d}}_{\text{max}}$ are bounds for the desired value.

\textit{Attitude Tracking}: The heading of UACV should point to the opponent, which means the Euler angle rather than the path angle should track the desired angle value in Eq. \eqref{desiredAngValue}. Therefore, when applying Eq. \eqref{loadCalculation} to calculate the load, all errors are replaced by $e_\chi=\chi^d-\theta$ and $e_\zeta=\zeta^\text{d}-\psi$.

\textit{Straight Flight}: The $\phi^\text{d}$ and $\chi^{\text{d}}$ are set as $0^\circ$.

\textit{Climb}: The $\phi^\text{d}$ and $\chi^{\text{d}}$ are set as zero and a positive value.

\textit{Somersault}: The $\phi^\text{d}$ is set as $0^\circ$ or $180^\circ$ according to whether the body is reversed or not. The desired angle of attack should be maintained as $\alpha^{\text{d}} = \alpha^{\text{full}}_{\text{max}}$.

\textit{Split-S}: This is a composite maneuver consisting of three steps. First, the UCAV reverses so that its normal direction points to the ground. Then, it pulls the load to maintain $\alpha^{\text{d}} = \alpha^{\text{full}}_{\text{max}}$ and decrease its height. Lastly, it changes the maneuver to straight flight after its pitch return to zero degree.

\textit{Yo-Yo}: This maneuver is used to exchange height and speed, and thus the tracking should have a vertical offset $\Delta h$. In this paper, the $\chi^\text{d}$ in Eq. \eqref{desiredAngValue} is rewritten as
\begin{equation}
	\chi^{\text{d}} = \arctan((-\Delta p_3\pm\Delta h)/\sqrt{\Delta p_1^2+\Delta p_2^2}),
\end{equation}
where the sign is set as positive for high yo-yo and negative for low yo-yo. Additionally, the offset is calculated as a certain ratio of the kinetic energy of the UCAV, which means
\begin{equation}
	\Delta h = k_h V^2/g.
\end{equation}

\subsection{Maneuver Decision Based on Decision Tree}
This paper constructs a DT including the baseline of tracking and three typical options. Self-protection is used to prevent the UCAV from falling into extreme conditions. We also design strategy for UCAV to escape from being chased from behind. Yo-yo is introduced to adjusts the approaching speed. The decision sequence is protection, escape, yo-yo, and lastly tracking. The three options are denoted with $SP$, $ES$, $YY$, which can be set as True or False to determine whether to use corresponding strategies. Details are as follows.

\textit{Self-Protection}: To prevent the UCAV from crashing, it selects the climb maneuver to increase its height if $-p_3 < h^{\text{protect}}$. Additionally, it selects the straight flight to recover its speed and maneuverability if $Ma<Ma^{\text{protect}}$.

\textit{Escape}: The UCAV escapes from disadvantage conditions. If $d<d^{\text{close}}$, $ATA>ATA^{\text{escape}}$, and $AA>AA^{\text{escape}}$, this paper assumes the UCAV is about to be chased from behind. Then, if $p_3<p_3^\text{o}$, it chooses somersault to escape from above. Otherwise, it selects split-s to escape from below.

\textit{Yo-Yo}: When $d>d^\text{close}$ and $ATA<ATA^\text{aim}$, this paper assumes the UCAV aims at the opponent from a distance. According to \cite{battleBook}, the UCAV chooses yo-yo when $AA^{\text{yo-yo}}_{\text{min}}<AA<AA^{\text{yo-yo}}_{\text{max}}$. Moreover, it selects a maneuver of high yo-yo to reduce speed and avoid overshoot when $V>V^\text{o}$, and low yo-yo to increase speed when $V<V^\text{o}$.

\textit{Tracking}: The UCAV needs to track its opponent. If $d<d^\text{close}$ and $ATA<ATA^\text{aim}$, it selects attitude tracking to aim its nose at the opponent and cause damage. Otherwise, it selects position tracking to approach the opponent.

\subsection{Maneuver Decision Based on DDQN}
During training, this paper let the agent control the blue UCAV, while the red one controlled with the DT strategy serves as the opponent. We modify the DDQN to train the agent to make maneuver decision for the dogfight, which is shown in Algorithm 1. Details are described as below.

\textit{State}: The state space of the agent is designed as 
\begin{equation}
	s_t = (p_3, V, \phi, \theta, \chi, \psi-\zeta^{\text{d}}, \xi - \zeta^{\text{d}}, HCA, ATA, AA, d, V^\text{o}).
\end{equation}
Then, we elaborate the conciseness of the state design. The absolute coordinate of the UCAV is omitted because only the relative position makes sense. Considering the isotropy of space, the relationship between two UCAVs when they rotate with the same angle has no changes. Therefore, the yaw is reduced to $\psi-\zeta^{\text{d}}$ and $\zeta - \zeta^{\text{d}}$. The position and attitude information of the opponent can be induced by $HCA$, $ATA$, $AA$, and $d$. Thus, they are not listed explicitly.

\textit{Action}: The action space is discrete and $|\mathcal{A}|=8$, where each one represents a BFM in Section III.B.

\textit{Reward}: The reward function in this paper is designed as
\begin{equation}
	\begin{aligned}
		r_t &= \lambda^{\text{f}}r^{\text{f}}_t + \lambda^{\text{d}}r^{\text{d}}_t + \lambda^{\text{a}}r^{\text{a}}_t, \\
		r^{\text{f}}_t &= \mathbb{I}(\text{red UCAV fails}) - \mathbb{I}(\text{blue UCAV fails}),\\		
		r^{\text{d}}_t &= (blood^{\text{red}}_{t-1} - blood^{\text{red}}_{t}) - (blood^{\text{blue}}_{t-1} - blood^{\text{blue}}_{t}),\\
		r^{\text{a}}_t &= 180 - ATA - AA.
	\end{aligned}
\end{equation}
Here, $\lambda$ means constant coefficients, $r^\text{s}_t$ represents the sparse reward determined by whether the agent or opponent fails, $r^\text{d}_t$ is the difference between the damage caused by two UCAVs, $r^{\text{a}}_t$ is the angle advantage. If the agent is defeated, damaged, or at a disadvantaged angle, it will receive a negative penalty. Otherwise, it will obtain a positive reward.

\textit{Transition}: Once the agent and its opponent select their BFM, the dynamical model updates for $\Delta T$ steps. If one of the UCAVs fails, then the simulation stops and the agent starts another episode. 

\begin{algorithm}[!t]
	\renewcommand{\algorithmicrequire}{\textbf{Input:}}
	\renewcommand{\algorithmicensure}{\textbf{Output:}}
	\caption{DDQN for UCAV Dogfight}  
	\begin{algorithmic}[1]
		\REQUIRE Initial network parameter $\eta$.  
		\ENSURE Optimized network parameter.
		\FOR {$\text{episode} = 1, 2, ...$} 
		\STATE Randomly initialize states of two UCAVs. 
		\FOR{$t = 0, 1, ...$}
		\STATE Sample maneuvers for red and blue UCAVs based on the DT in Section III-C and Eq. \eqref{epsilonGreedy}, respectively.
		\STATE Set $done_t$ as False.
		\FOR{$t'=1,2,...,\Delta T$}
		\STATE Calculate inputs based on Section III-A and B.
		\STATE Update dynamics based on Section II-A
		\IF{One of UCAVs fails according to Section II-C}
		\STATE Set $done_t$ as True, and break.
		\ENDIF
		\ENDFOR
		
		\STATE Calculate the next state $s_{t+1}$ and reward $r_t$.
		\STATE Store the trajectory $(s_t, a_t, s_{t+1}, r_t, done_t)$ into $\mathcal{D}$.
		\STATE Uniformly sample trajectories from $\mathcal{D}$.
		\STATE Update $\eta$ with the loss function of Eq. \eqref{lossF}.
		\STATE Update parameters of the target network with $\hat{\eta}\leftarrow\eta$ for every $\widetilde{C}$ training steps.
		\ENDFOR
		\ENDFOR
	\end{algorithmic} 
\end{algorithm}

\section{Simulation}

\subsection{Simulation Setup}
This paper selects F16s as the UCAV model, where all parameters follow \cite{F16}. PID for the low-level control are set as $\PIDcoe_{Ma}=[10, 0, 0]^T$, $\PIDcoe_{\alpha}=[0.8, 1.8, 10]^T$, $\PIDcoe_{\phi}=[0.07, 0, 0]^T$, and $\PIDcoe_{\beta}=[12, 0, 4]^T$. Here, roll and sidelip channels reduce to P and PD control respectively, because their trims are zero. 

Parameters of tracking are set as $k_\zeta=k_\chi=0.02$, $k_n^\alpha=4$. 
Bounds of $\alpha$ 
are set as $\alpha^{\text{d}}_{\text{min}}=-4^\circ$ and $\alpha^{\text{d}}_{\text{max}}=20^\circ$. Moreover, we set $\alpha_{\text{max}}^{\text{full}}=30^\circ$ to give escape BFM higher maneuverability, and $k_h=0.1$ for yo-yo.

The protection bounds are set as $h^{\text{protect}}=1000$ m and $Ma^{\text{protect}}=0.3$. In addition, this paper set $d^\text{close}=3000$ m and $ATA^\text{aim}=30^\circ$ for tracking, $ATA^\text{escape}=AA^\text{escape}=120^\circ$ for escape, and $AA^{\text{yo-yo}}_{\text{min}}=30^\circ$, $AA^{\text{yo-yo}}_{\text{max}}=60^\circ$ for yo-yo \cite{battleBook}. The desired path pitch for climb is $20^\circ$.

When applying the DRL algorithm, the simulation step is set as $0.01$ s. The decision period is set as $1$ s, namely $\Delta T = 100$. The initial position, speed belong to $[-3,3]\times[-3,3]\times[-3,-8]$ km and $[0.3, 0.9]$ mach. Initial pitch and roll are set as zero, while yaw belongs to $[-180, 180]$ degree. Each episode lasts for $300$ time steps. As for the setting of DDQN, this paper set $|\mathcal{D}|=1\times10^5$, $\epsilon=0.95$, $\gamma=0.95$, and $\widetilde{C}=512$. The sparse reward is set as $\lambda^\text{f}=20$. Dense ones are set as $\lambda^{\text{d}}=1$ and $\lambda^{\text{a}}=1/180$, which normalize each term to $[-1,1]$.

Moreover, \cite{DARPA} indicates that wide shallow network is more suitable for DRL algorithm. Therefore, this paper adopts networks with only two hidden layers, each containing $512$ and $256$ nodes. Adam is applied as the optimizer, where the learning rate and batch size are set as $1\times 10^{-4}$ and $512$. 

\begin{figure}[!t]
	\centering
	\includegraphics[]{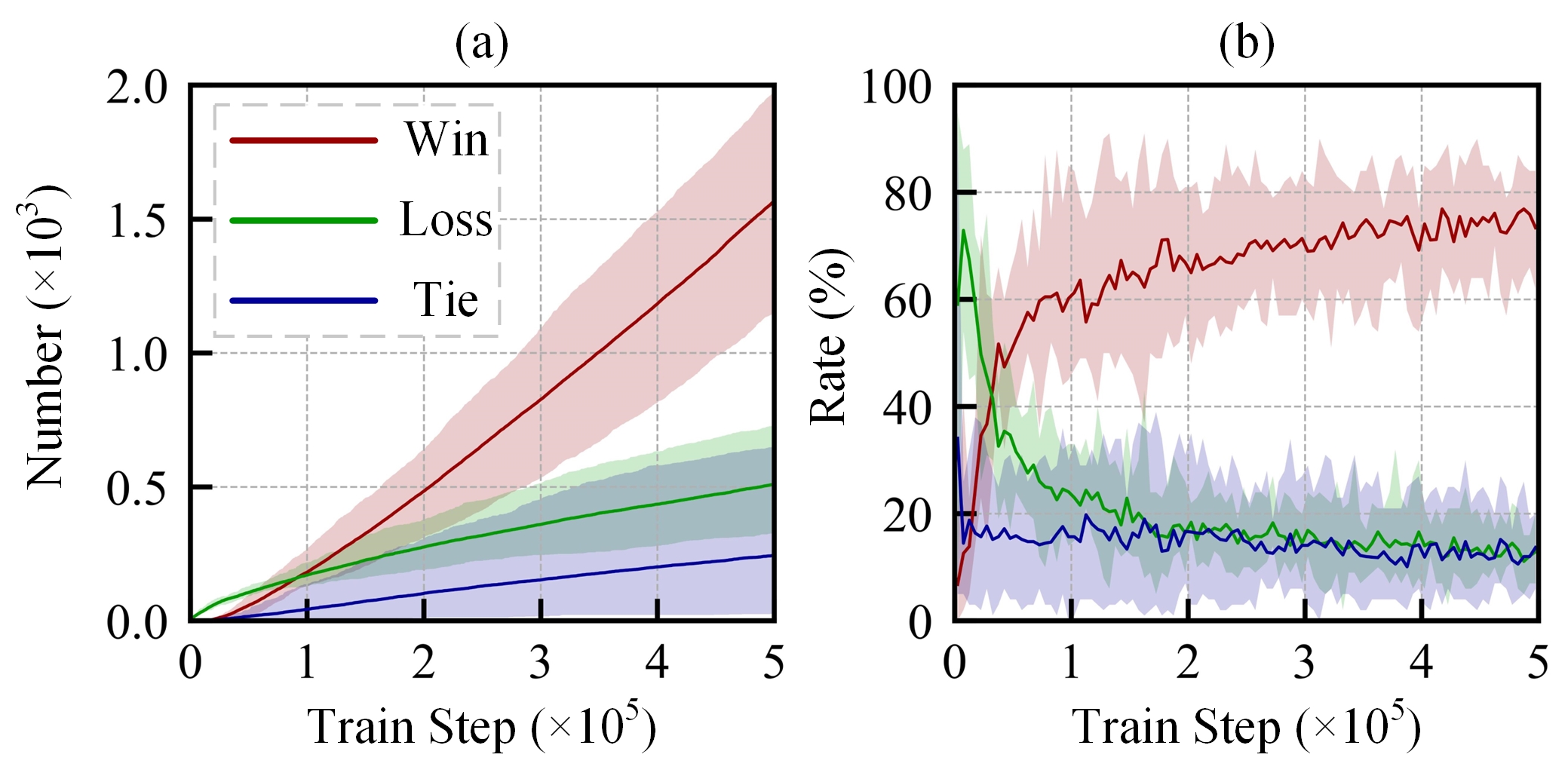}
	\caption{(a) Accumulated number and (b) rate for win, loss, tie among the training process. The solid lines mean the average calculated over 10 trials, while the shadow ones represent the maximum and the minimum.}
\end{figure}

\subsection{Training Process}
Ten trails containing $5\times10^5$ training steps are tested. The time consumption of each trail is approximately $11$ hours, tested on a computer with i9-13900K CPU. The accumulated number of win, loss, and tie for the agent is shown in Fig. 3 (a). 
Then, we randomly initialize $400$ circumstance to test the rate of win, loss, and tie of agents against the DT, which is shown in Fig. 3 (b). In general, all trails converges and agents outperform the DT after around $5\times10^4$ steps of training. However, similar to numerous previous works, the training process of DRL has significant randomness and fluctuation. Some trails could achieve the highest success rate of 90\% under $100$ randomly initialize testing circumstances, while others could only achieve 80\%. Therefore, this paper selects the best-performed agent for the following analysis.

\subsection{Robustness Analysis}
The strategy of opponents during the training process is a full DT, which may vary in real dogfight conditions. Therefore, this section test the robustness of the agent facing with different unseen opponents. We randomly initialize $400$ circumstances. The performance of the best-performed agent facing eight strategies with different settings of protection, escape, and yo-yo are tested, which is shown in Table I.

\begin{table}[!t]
	\caption{Rate of win, loss and tie of the best-performed agent facing with different unseen opponents.}
	\centering
	\begin{tabular}{|c|ccc|ccc|}
		\hline
		Strategy & $SP$ & $ES$ & $YY$ & Win & Loss & Tie \\ \hline
		1    &    &    &    & 54.00  & 30.50   & 15.50    \\ 
		2    & \checkmark    &    &    & 70.75  & 15.25   & 14.00    \\
		3    &    & \checkmark    &    & 56.50  & 30.50   & 13.00    \\ 
		4    &    &    & \checkmark    & 53.00  & 31.50   & 15.50     \\ 
		5    &    & \checkmark  & \checkmark  & 55.75  & 31.25   & 13.50    \\ 
		6    & \checkmark  &    & \checkmark  & 70.00  & 15.25    & 14.75     \\ 
		7    & \checkmark  & \checkmark  &    & 85.50  & 9.50   & 5.00    \\ 
		8   & \checkmark  & \checkmark  & \checkmark  & 85.75  & 9.25    & 5.00     \\ \hline
	\end{tabular}
\end{table}

In general, the agent can achieve positive results when facing all opponent. Nevertheless, its performance dramatically declined within some conditions, which indicate the need of further study on increasing DRL robustness. 

Additionally, statistical results indicate following phenomena. Firstly, whether the opponent apply yo-yo has little impact on the agent performance. That is because the distance between two UCAVs are almost always lower than $d^\text{close}=3000$ m, which means the opponent rarely conduct yo-yo BFM. Then, the protection mechanism has the most important impact on the agent performance. Specifically, this paper claims that the agent learns how to leverage the height protection to defeat the opponent. Furthermore, the escape strategy can bring improvement to the agent only when the opponent adopts protection simultaneously. 


\subsection{Case Analysis: Double Loop}
To illustrate how the agent trained by the proposed method defeat the opponent in detail, this paper constructs two typical cases.
The initial condition of the first case is set as $\pos^{\text{red}}=[0, -2000, -5000]$, $\pos^{\text{red}}=[0, 2000, -5000]$, $\psi^{\text{red}}=0$,  $\psi^{\text{blue}}=180$, $Ma^{\text{red}}=Ma^{\text{blue}}=0.9$. This case lasts for $145$ s, and forms a typical double-loop dogfight condition shown in Fig. 4. 

\begin{figure}[!t]
	\centering
	\includegraphics[]{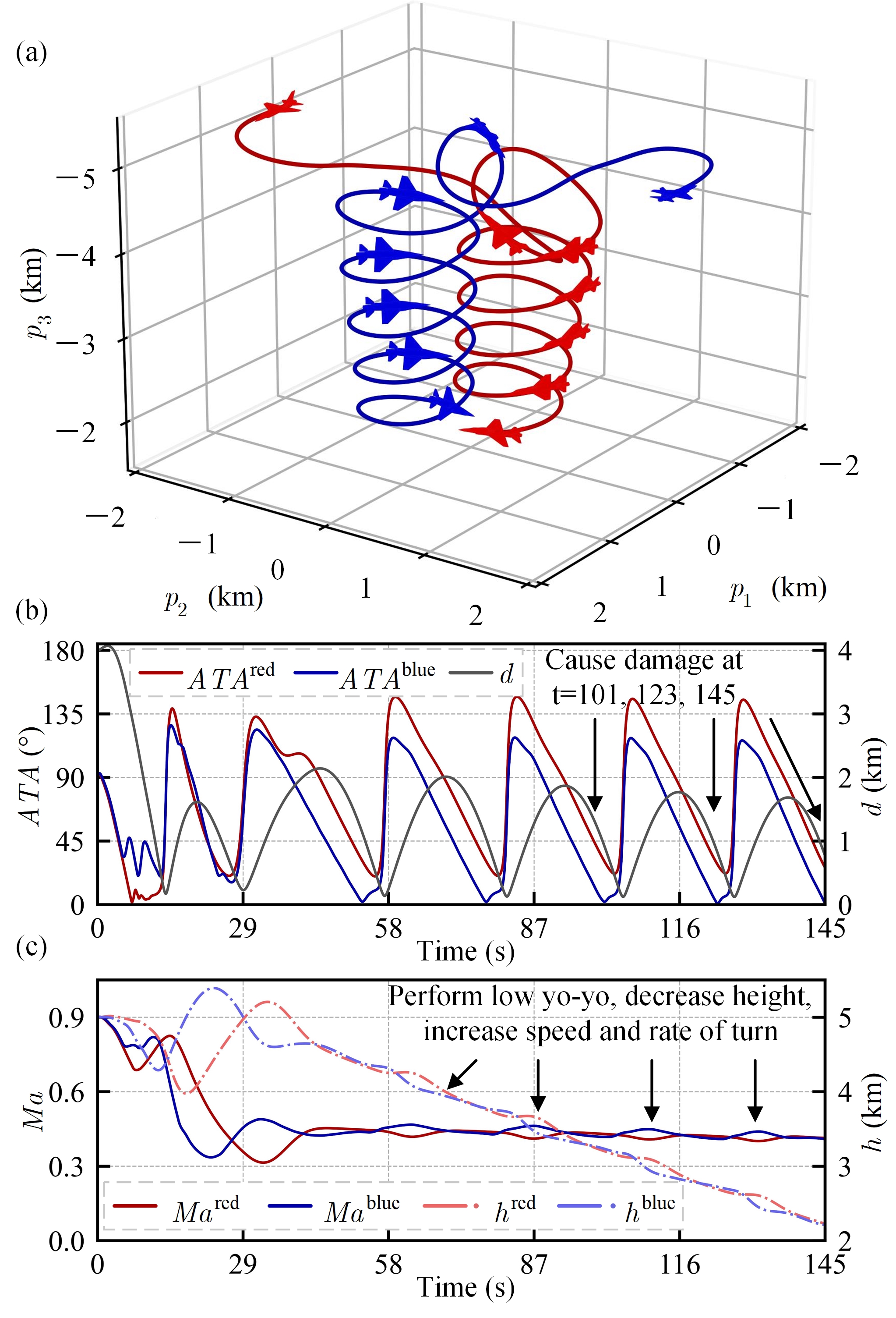}
	\caption{(a) 3D trajectories, and response of (b) ATA, distance, (c) mach, height of two UCAVs in the first case with double-loop dogfight condition.}
\end{figure}

At the very beginning, the agent performs low yo-yo to decrease its height, and then somersault (at $t=12$) to gain being damaged during merging. Compared with tracking, somersault can provide the agent with higher maneuverability, and thus build advantage after merging. At approximately $t=37$ s, the agent has built sufficient angle advantage, and then it begins to perform tracking BFM. Both UCAVs track each other, and therefore they form a double loop. Furthermore, the agent always chooses low yo-yo after each merging. It uses height to exchange kinetic energy, and thus gains higher turn rate to build angle advantage. After two rounds of engagements, it begins to cause damage to the opponent at $t = 101, 123, 145$ s, and wins the dogfight without being attacked.
Therefore, this case illustrates that, the agent trained with DRL can perform higher maneuverability compared with DT algorithm. 

\subsection{Case Analysis: Dive and Chase}

The second case illustrates a learned strategy named as "Dive and Chase" by which the agent leverages the protection and escape mechanism of the opponent. The initial mach is set as 0.8, where other conditions are the same with the first case. Fig. 5 shows trajectories and response of this case.

\begin{figure}[!t]
	\centering
	\includegraphics[]{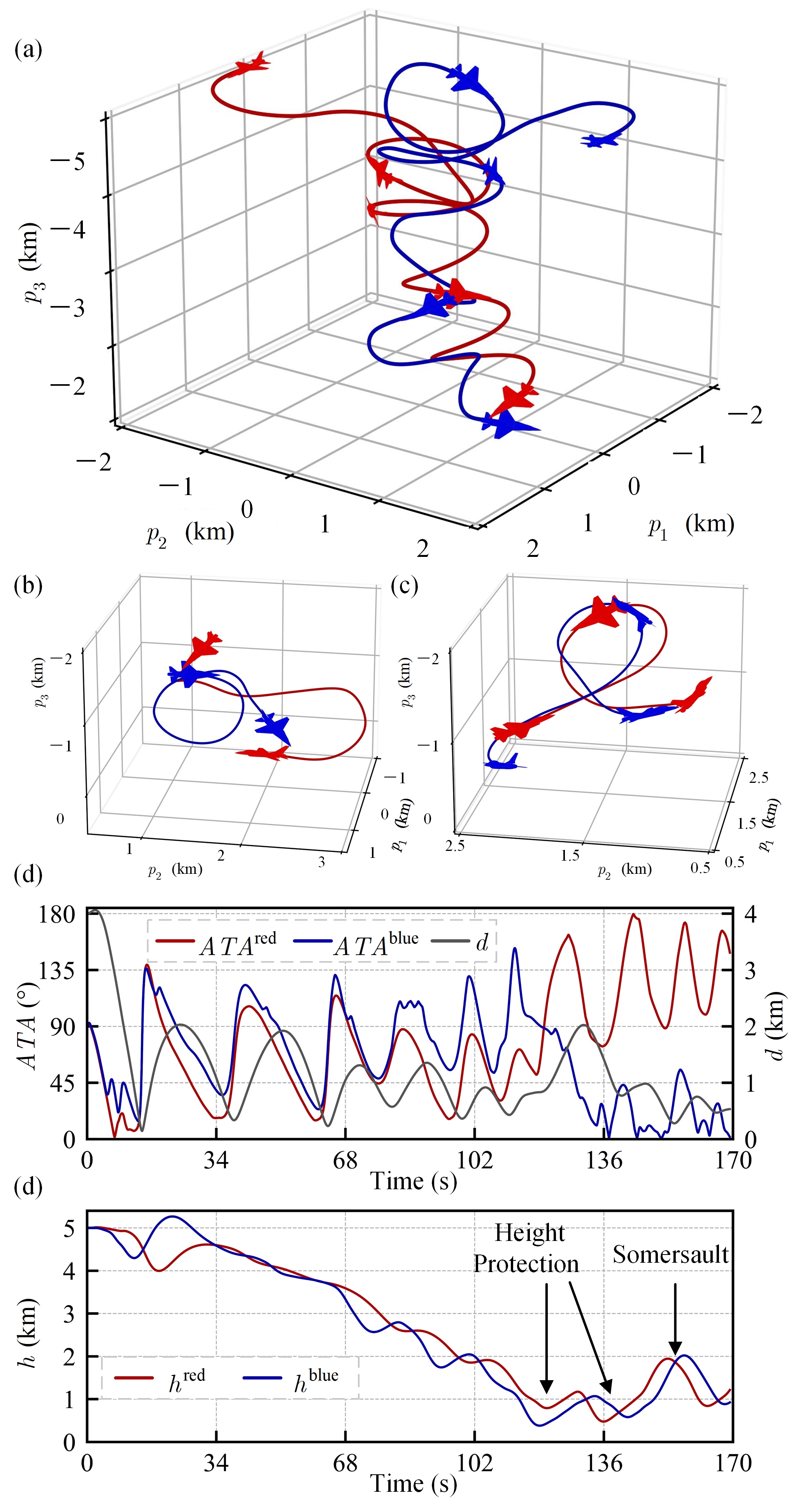}
	\caption{3D trajectories during (a) $[0, 110)$, (b) $[110, 140)$, (c) $[140, 169]$ s, and response of (d) ATA, distance, and (c) height in the second case.}
\end{figure}

During the period of $[0, 11000]$ s, the agent mixes numerous low yo-yo and split-s. This behavior causes the height of two UCAVs rapidly drop. After $t= 110$ s, the agent performs a compound maneuver of "spli-s, somersault, tracking" at a low altitude. This behavior triggers the height protection of the opponent, and builds angle advantage for the agent. After $t=130$ s, the opponent tracks the agent, but falls below $h^\text{protect}=1000$ m for the second time. Then, the agent tracks its opponent who climbs due to height protection, and builds a significant angle advantage. The opponent then performs somersault to escape from being chased. This paper claims that, the agent can recognize the high maneuverability escape BFM through the state. Hence, the agent abandons tracking and selects somersault at the same time, by which it can maintain its angle advantage even its opponent intends to escape, as shown in Fig. 5 (c). At last, the agent wins. The emergence of "Dive and Chase" strategy during the DRL training process indicates that, the agent can defeat opponents built by expertise from a tactical perspective. 




\section {Conclusion}
In this paper, we propose a DRL-based maneuver decision frame for the dogfight. A four-channel low-level control law is firstly designed, and then eight commonly used BFMs are then constructed. A DT consisting of four typical strategies is built. Lastly, we introduce DDQN for the UCAV maneuver decision. Our simulation result shows that, DRL can enormously outperform the conventional DT method from the perspective of both maneuverability and tactic.

\bibliographystyle{IEEEtran}
\balance
\bibliography{refs}

\end{document}